\documentclass[a4paper]{article}

\usepackage{authblk}
\usepackage[utf8]{inputenc}
\usepackage[T1]{fontenc}
\usepackage{pslatex}
\usepackage[english]{babel}
\usepackage{fancyhdr}
\usepackage{hyperref}
\usepackage{amsmath,amssymb,amsfonts}
\usepackage{graphicx}
\usepackage{textcomp}
\usepackage{xcolor}
\usepackage{algorithm}
\usepackage[noend]{algpseudocode}
\usepackage{url}
\usepackage{hyperref}
\usepackage{microtype}
\usepackage{subcaption}
\usepackage{tikz}
\usetikzlibrary{positioning}
\usetikzlibrary{decorations.pathreplacing}

\newcommand{\N}{\mathbb{N}}
\newcommand{\F}{\mathbb{F}}
\newcommand{\Z}{\mathbb{Z}}

\newtheorem{lemma}{Lemma}
\newtheorem{problem}{Problem}

\definecolor{darkpastelgreen}{rgb}{0.01, 0.75, 0.24}

\providecommand{\keywords}[1]{\textbf{\textit{Keywords }} #1}

\begin{document}

\title{The Influence of Local Search over Genetic Algorithms with Balanced Representations}

\author[1]{Luca Manzoni}
\author[2]{Luca Mariot}
\author[3]{Eva Tuba}
	
\affil[1]{{\normalsize Dipartimento di Matematica e Geoscienze, Universit\`{a} degli Studi di Trieste, Via Valerio 12/1, 34127 Trieste, Italy} \\
	
	{\small \texttt{lmanzoni@units.it}}}

\affil[2]{{\normalsize Digital Security Group, Radboud University, Postbus 9010, 6500 GL Nijmegen, The Netherlands} \\
	
	{\small \texttt{luca.mariot@ru.nl}}}
	
\affil[4]{{\small Faculty of Informatics and Computing, Singidunum University, Danijelova 32, 11000 Belgrade, Serbia}

     {\small \texttt{etuba@ieee.org}}}
	
\maketitle

\begin{abstract}
We continue the study of Genetic Algorithms (GA) on combinatorial optimization problems where the candidate solutions need to satisfy a balancedness constraint. It has been observed that the reduction of the search space size granted by ad-hoc crossover and mutation operators does not usually translate to a substantial improvement of the GA performances. There is still no clear explanation of this phenomenon, although it is suspected that a balanced representation might yield a more irregular fitness landscape, where it could be more difficult for GA to converge to a global optimum. In this paper, we investigate this issue by adding a local search step to a GA with balanced operators, and use it to evolve highly nonlinear balanced Boolean functions. In particular, we organize our experiments around two research questions, namely if local search (1) improves the convergence speed of GA, and (2) decreases the population diversity. Surprisingly, while our results answer affirmatively the first question, they also show that adding local search actually \emph{increases} the diversity among the individuals in the population. We link these findings to some recent results on fitness landscape analysis for problems on Boolean functions. 
\end{abstract}


\section{Introduction}
\label{sec:intro}
There exist two common approaches for constraint handling in the literature of Genetic Algorithms (GA): incorporate a penalty factor in the fitness function that punishes deviations from the desired constraints, or employ ad-hoc representations and variation operators. Penalty factors are fairly simple to implement and can be employed virtually in any optimization problem, once a suitable notion of distance from the required constraints has been defined. However, penalty factors can also be wasteful, since a GA may spend a great amount of fitness evaluations to satisfy them, driving the search effort away from the main optimization objective. The second approach requires designing suitable crossover and mutation operators, so that feasible parents produce feasible offspring. This makes the GA explore a smaller search space, which in principle should lead to better performances, since the fitness budget is entirely used to evolve feasible solutions only. 

In this work, we focus on the second approach for handling \emph{balancedness} constraints, namely when the binary representation of the candidate solutions must have a fixed number of ones. Such a constraint is relevant in several optimization problems related to cryptography, coding theory and combinatorial designs. To the best of our knowledge, Lucasius and Kateman~\cite{lucasius92} were the first to investigate balancedness-preserving crossover operators in GA, applying them to the subset selection problem. Millan et al.~\cite{millan98} used GA to evolve balanced Boolean functions with good cryptographic properties such as high nonlinearity and low deviation from correlation immunity. To this end, the authors devised a counter-based crossover operator that preserved the balancedness of the parent Boolean functions. Balanced crossover operators have also been designed for other optimization problems such as portfolio optimization~\cite{chen2006,chen2009} and multiobjective $k$-subset selection~\cite{meinl2009}. Further extensions of this approach include the design of balancedness-preserving operators for non-binary candidate solutions with non-binary representations~\cite{mariot15,mariot17} or for matrix-based representations where each column needs to be balanced~\cite{mariot18}.

More recently, we carried out in~\cite{manzoni20} a rigorous statistical investigation of three balanced crossover operators against different optimization problems related to cryptography and combinatorial designs. We found that balanced operators indeed give an advantage to GA over a classic one-point crossover coupled with a penalty factor. Hence, these results seem to confirm the aforementioned principle that reducing the search space by means of ad-hoc variation operators improves the GA performance. Nonetheless, the improvement is not substantial and does not scale well with respect to the problem size. This is especially evident when comparing a GA based on balanced crossover operators with other metaheuristics such as Genetic Programming (GP). In general, it has been observed that GP converges more easily to an optimal solution than GA on problems where balanced solutions are sought~\cite{picek16,mariot17,mariot18}.

Clearly, the use of ad-hoc operators can change the fitness landscape of a particular optimization problem. Indeed, one of the possible explanations for the meagre improvement of GA when using balanced crossover operators is that the resulting fitness landscape becomes more irregular. Hence, although searching a smaller space of feasible solutions, the GA could get stuck more easily on local optima. We started to investigate this hypothesis in~\cite{manzoni21} by considering an \emph{adaptive bias} strategy where the counter-based crossover of~\cite{millan98} is allowed to produce partially unbalanced Boolean functions. The rationale is that, by slightly enlarging the search space, the GA might escape more easily from local optima, thus improving its explorability. Yet, the results showed that even this strategy provides only a marginal improvement in the GA performances.

In this paper, we further investigates the scarce improvement of GA with balanced crossover operators by augmenting them with a \emph{local search step}. In particular, we consider the evolution of highly nonlinear balanced Boolean functions as an underlying optimization problem, for which an efficient local search move has already been developed in~\cite{millan99}. We perform an experimental evaluation of the three balanced crossover operators in~\cite{manzoni20} by combining them with three variants of local search. The first variant is the baseline GA where no local search is performed. The second variant applies only a single step of local search on a new offspring individual created through balanced crossover and mutation. The third variant, finally, is a steepest ascent strategy, which performs local search on an offspring individual until a local optimum is reached. The experiments are performed for Boolean functions of $6 \le n \le 9$ variables.

To assess the influence that local search has over the GA performances, we consider two research questions. The first one is whether local search improves the convergence speed of GA to a local optimum. As expected, the answer given by our experimental results is positive, especially for the third variant employing the steepest ascent strategy. On the other hand, the second research question is whether the use of local search decreases the diversity in the population, as measured by the pairwise Hamming distance. Indeed, a natural hypothesis for the scarce improvement of GA performances when using balanced crossover operators is that the solutions in the population become too similar, determining a premature convergence to a local optimum. Therefore, one would expect that such a phenomenon is magnified by augmenting the GA with a local search step. Surprisingly, our results indicate that the use of local search actually \emph{increases} the population diversity. We discuss this interesting finding by linking it to a recent work on fitness landscape analysis for problems related to cryptographic Boolean functions~\cite{jakobovic21}. In particular, the fact that the individuals in the population tend to be quite different among each other seem to indicate that the fitness landscape of balanced Boolean functions is characterized by many isolated local optima. This in turn suggests that a possible way to improve the GA performances is to use a different initialization strategy that the usual one where candidate solutions are generated uniformly at random.

The rest of this paper is organized as follows. Section~\ref{sec:bg} covers all background definitions related to balanced crossover operators and Boolean functions that the contributions of this paper are based upon. Section~\ref{sec:ls} defines the optimization problem of evolving highly nonlinear balanced Boolean function, and describes the local search algorithm used as a further optimization step after balanced crossover and mutation. Section~\ref{sec:exp} presents the experimental evaluation of our approach, discussing the experimental settings adopted and the obtained results. Finally, Section~\ref{sec:outro} concludes the paper by summarizing the main findings and pointing out directions for further research on the topic.

\section{Background}
\label{sec:bg}
In this section, we first describe the three balanced crossover operators introduced in~\cite{manzoni20}, which we will use in our investigation. Next, we recall the basic notions related to Boolean functions and their cryptographic properties, that will be the basis of the underlying optimization problem for our experiments.

As a general notation, in what follows we denote by $\F_2 = \{0,1\}$ the finite field with two elements, and $\F_2^n$ is the set of all $n$-bit strings, which is endowed with a vector space structure. In particular, the sum of two vectors $x,y \in \F_2^n$ corresponds to their bitwise XOR $x \oplus y$, while multiplication of $x \in \F_2^n$ by a scalar $a \in \F_2$ amounts to computing the logical AND of $a$ with each cooordinate of $x$. The \emph{scalar product} of two vectors $x, y \in \F_2^n$ is defined as $\bigoplus_{i=1}^n x_iy_i$, i.e. the XOR of all bitwise AND of the two vectors. Given $[n] = \{1,\cdots, n\}$ for all $n \in \N$, the \emph{Hamming distance} of $x,y \in \F_2^n$ is defined as $d_H(x,y) = |\{ i \in [n]: x_i \neq y_i \}|$, i.e. the number of coordinates where $x$ and $y$ differ. The \emph{Hamming weight} of a vector $x \in \F_2^n$, denoted by $w_H(x)$, is the Hamming distance of $x$ from the null vector $\underbar{0}$, or equivalently the number of ones in $x$. The number of binary strings with a fixed Hamming weight $k \in [n]$ is the binomial coefficient $\binom{n}{k}$, since it is equivalent to the number of $k$-subsets of $[n]$, when one interprets a vector $x \in \F_2^n$ as the characteristic function of a subset.

\subsection{Balanced Crossover Operators}
\label{subsec:balcros}
We start by giving a brief description of the three balanced crossover operators that we will use in our experiments. Further details about them and their pseudocode can be found in our previous paper~\cite{manzoni20}. In the remainder of this paper, we assume that the Hamming weight that we want to preserve is exactly half of the string length, i.e. the individuals in the population have an equal number of zeros and ones in their representation.

\subsubsection{Counter-Based Crossover}
\label{subsec:cbcx}
The first operator employs two counters $cnt_0$ and $cnt_1$ to keep track respectively of how many zeros and ones the child individual has during the crossover process. Specifically, given two parent bitstrings $p_1,p_2 \in \F_2^{2m}$ such that $w_H(p_1) = w_H(p_2) = m$, a child chromosome $c \in \F_2^{2m}$ is obtained by randomly copying either the $i$-th bit of $p_1$ or $p_2$ with uniform probability, for each position $i \in [2m]$. Then, $cnt_0$ or $cnt_1$ is incremented depending on the value copied in the child. When one of the two counters reaches the threshold weight $m$, the remaining positions in the child are filled with the complementary value.

A natural question about this crossover operator is whether setting the last bits to a fixed value to preserve balancedness does not introduce a bias towards certain solutions in the search space. We considered this issue in our previous work~\cite{manzoni20}, by comparing the basic "left-to-right" version of the operator described above with another one that randomly shuffles the order of the positions to be copied in the child chromosome. Results showed that in most cases there is no significant difference among the two variants, while in certain instances the shuffling strategy fares even worse than the basic "left-to-right" version. Hence, we used the latter for the experiments of this paper. 

\subsubsection{Zero-Length Crossover}
\label{subsec:zlcx}
The second crossover operator considered in our investigation is based on a different representation of the candidate solutions, namely their \emph{zero-length} encoding. Formally, given a $n$-bit string $x$ with $n=2m$, the zero-length encoding of $x$ is a vector $r$  of length $m+1$ where each coordinate $r_i$ represents the number of consecutive zeros (or equivalently, the run length of zeros) between two consecutive ones.

To correctly represent a balanced bitstring, the values in the zero-length encoding vector must sum to $m$. Sticking to our previous example, the zero-length encodings of $p_1 = (0,1,0,1,0,1,1,0)$ and $p_2 = (1,0,0,0,1,0,1,1)$ are respectively $r_1 = (1, 1, 1, 0, 1)$ and $r_2 = (0, 3, 1, 0, 0)$. At each position the zero-length crossover randomly copies the zero-length value of the first or second parent with uniform probability. An accumulator variable is used to represent the partial sums of the zeros' run lengths in the offspring chromosome. If the threshold value $m$ is reached, the remaining positions of the offspring's zero-length vector are filled with zeros; thus, the bitstring representation will only contain ones in the last positions. Otherwise, the last coordinate of the zero-length vector is filled with the value that balances the sum to $m$; accordingly, the bitstring representation of the offspring will contain only zeros in the last positions.

\subsubsection{Map-of-Ones Crossover}
\label{subsec:mocx}
The third crossover considered in our experiments leverages on an integer-based representation of the candidate solutions. In particular, the map-of-ones is simply the vector that indicates the positions of the ones in a bitstring. Using our examples above, the map of ones for $p_1 = (0,1,0,1,0,1,1,0)$ and $p_2 = (1,0,0,0,1,0,1,1)$ are $b_1 = (2,4,6,7)$ and $b_2 = (1,5,7,8)$, respectively. Similarly to the previous two operators, the map-of-ones crossover works coordinate-wise by randomly copying either the value of the first or second parent's zero-length vector in the child chromosome. The only constraint that is enforced is that the map of ones of the child chromosome cannot have duplicate values, something that can occur if the bitstrings of the two parents have value one in the same position. For this reason, the crossover first computes a list of common positions between the two parents, and then cat any step checks whether the selected value has already been inserted before in the child or not. If this is the case, then the value from the other parent is copied instead.

\subsection{Boolean Functions}
\label{subsec:bf}
We now describe the essential notions related to the optimization problem underlying our experiments on local search. A Boolean function of $f: \F_2^n \to \F_2$ is a mapping $f: \F_2^n \to \F_2$, i.e. a function that associates to each $n$-bit vector a single output bit, $0$ or $1$. The most common way to represent such a function is via its \emph{truth table}: assuming that the vectors of $\F_2^n$ are lexicographically ordered, the truth table of $f$ is the $2^n$-bit vector
\begin{displaymath}
\Omega_f = (f(0,\cdots,0), f(0,\cdots,1), \cdots, f(1,\cdots,1)) \enspace ,
\end{displaymath}
i.e. the vector that specifies the output value $f(x)$ for each possible input vector $x \in \F_2^n$. A fundamental criterion for Boolean functions used in stream ciphers is that the truth table must be a balanced string, i.e. $w_H(f) = 2^{n-1}$, to resist basic statistical attacks.

Another way to uniquely represent a Boolean function commonly used in cryptography is the Walsh transform. Formally, the Walsh transform of $f: \F_2^n \to \F_2$ is the map $W_f: \F_2^n \to \Z$ defined as:
\begin{equation}
\label{eq:wht}
W_f(a) = \sum_{x \in \F_2^n} (-1)^{f(x) \oplus a \cdot x} \enspace ,
\end{equation}
for all $a \in \F_2^n$. The coefficient $W_f(a)$ measures the correlation between $f$ and the linear function defined by the scalar product $a \cdot x$. A second important property for Boolean functions used in symmetric cryptography is their nonlinearity, which is defined as:
\begin{equation}
\label{eq:nl}
nl(f) = 2^{n-1} - \frac{1}{2}\max_{a \in \F_2^n} \{|W_f(a)|\} \enspace .
\end{equation}
We refer the reader to~\cite{carlet21} for further cryptographic implications and bounds related to the nonlinearity property. Here, we just limit ourselves to specify that the nonlinearity should be as high as possible. Taking into account also the balancedness property mentioned above, this gives rise to the following optimization problem:
\begin{problem}
\label{pb:opt}
Let $n \in \N$. Find a $n$-variable Boolean function $f: \F_2^n \to \F_2$ that is balanced and has maximum nonlinearity, as measured by the fitness function $fit(f) = nl(f)$. 
\end{problem}

Remark that it is still an open question to determine the maximum nonlinearity value attainable by a balanced Boolean function for $n>7$ variables~\cite{carlet21}. We will tackle Problem~\ref{pb:opt} in the experimental part of the paper using various combinations of balanced GA and local search.

\section{Local Search of Boolean Functions}
\label{sec:ls}
To perform local search, the first step is to define an \emph{elementary move} between two candidate solutions. This further subsumes the notion of a topology over the search space, in order to give a precise meaning to the \emph{neighborhood} of a solution. In our case, since we are dealing with fixed-length binary strings to represent the truth tables of Boolean functions, the most obvious choice is to adopt the topology induced by the Hamming distance. Therefore, the neighborhood of a candidate a solution $f: \F_2^n \to \F_2$ represented by its truth table $\Omega_f \in \F_2^{2^n}$ would be the set of all truth tables at Hamming distance $1$ from $\Omega_f$. Hence, the elementary move from $f$ to a neighboring solution $f'$ would be obtained by complementing a single bit in $\Omega_f$. However, such a move would break the balancedness constraint, since the Hamming weight would change by $\pm 1$. Hence, similarly to the mutation operator employed in our previous paper~\cite{manzoni20}, we consider the \emph{swap} between two different values in $\Omega_f$ as an elementary move for our local search procedure. In this way, the Hamming weight of the new candidate solution will still be $2^{n-1}$.

Concerning the Walsh transform, a single swap in the truth table of $f$ induces a change $\Delta(a) \in \{-2, 0, +2\}$ for each coefficient $a \in \F_2^n$, that can be computed with the following result proved in~\cite{millan99}:
\begin{lemma}
\label{lm:hc}
Let $f: \F_2^n \to \F_2$ be a $n$-variable Boolean function, and assume that $y,z \in \F_2^n$ are such that $f(y) \neq f(z)$. Then, for each $a \in \F_2^n$, it holds that:
\begin{equation}
\label{eq:delta}
\Delta(a) = (-1)^{f(y)} [(-1)^{a\cdot z} - (-1)^{a \cdot y}] + (-1)^{f(z)} [(-1)^{a \cdot y} - (-1)^{a \cdot z}] \enspace .
\end{equation}
\end{lemma}
Consequently, there is no need to recompute the Walsh transform from scratch when swapping two values in the truth table of $f$. Using Lemma~\ref{lm:hc}, each coefficient can be updated from the old one as $W_{f'}(a) = W_f(a) + \Delta(a)$. This allows one to efficiently explore the neighborhood of a given function, since in this way the fitness of a single swap can be evaluated in linear time with respect to the length of the function's table. On the other hand, recomputation from scratch would entail a quadratic complexity by using the \emph{fast Walsh transform} algorithm~\cite{carlet21}, which is the one employed by the GA to evaluate the fitness of a new individual created through crossover and mutation.

In summary, a single iteration of the GA combined with a local search step works as follows:
\begin{enumerate}
\item Select a pair of parents $p_1$, $p_2$ from the population.
\item Apply crossover and mutation to obtain a new balanced individual $c$.
\item Evaluate the fitness of $c$ by computing the Walsh transform in Eq.~\ref{eq:wht} using the fast algorithm~\cite{carlet21}.
\item Apply one or more steps of local search to $c$ as follows:
\begin{enumerate}
    \item Generate the \emph{2-Improvement set} of $c$, i.e. finds all swaps in $c$ such that the nonlinearity increases by $2$. Use Eq.~\ref{eq:delta} to efficiently update the Walsh transform for each swap
    \item Pick a swap in the 2-Improvement set and apply it to $c$, updating the fitness value as $W_{c'}(a) = W_c(a) + \Delta(a)$ for all $a \in \F_2^n$.
\end{enumerate}
\end{enumerate}
Since each swap in the improvement set increases the nonlinearity by $2$, there is no ground to drive the selection. In our experiments, we pick the first generated swap. This is similar to the strategy adopted in~\cite{jakobovic21} where local search was used to create the \emph{Local Optima Network} of the search space of Boolean functions.

\section{Experiments}
\label{sec:exp}
As discussed in the Introduction, our aim is to assess the influence of local search as a further optimization step in the loop of a GA with balanced crossover. To this end, we consider the following two research questions:
\begin{itemize}
\item {\bfseries RQ1}: does local search improve the convergence speed of GA, i.e. does it allow to reach a local optimum in less fitness evaluations?
\item {\bfseries RQ2}: does local search decrease the diversity of the GA population?
\end{itemize}
Remark that we deliberately excluded any research question pertaining the improvement of the best fitness. Indeed, it has already been remarked that balanced GA usually have a lower performance than other metaheuristics on combinatorial optimization problems such as Problem~\ref{pb:opt}. Moreover, in~\cite{manzoni21} we observed that augmenting a balanced GA with a partially unbalanced crossover strategy does not improve significantly the best fitness. Considering also the evidence gathered in~\cite{millan98} where a balanced GA combined with hill climbing was used, our hypothesis is that local search step does not make a significant difference as well. As we will show in the next sections, this hypothesis was experimentally confirmed.

Nevertheless, it is reasonable to expect that adding local search in the loop may help GA to converge more quickly toward a local optimum, which motivates {\bfseries RQ1}. Furthermore, crossover tends to exploit the genetic information of the current population, producing offspring individuals that resemble their parents, and thus decreasing the population diversity. Therefore, one may also expect that a local search step would magnify this effect, by tweaking the candidate solutions toward the nearest local optimum. This argument motivates {\bfseries RQ2}.

In what follows, we describe the experimental settings used to investigate our research questions and the results obtained from our experiments. 

\subsection{Experimental Setting}
\label{subsec:exp-set}
For our experiments, we tested three variants of local search, namely: 
\begin{itemize}
\item $LS0$: No local search, which corresponds to the basic balanced GA.
\item $LS1$: Single-step local search, where only a single swap is performed on a new individual.
\item $LS2$: Steepest ascent local search, with swaps performed until a local optimum is reached.
\end{itemize}

We considered counter-based ($CX1$), zero-length ($CX2$) and map-of-ones ($CX3$) crossover. As for mutation, we adopted the simple swap-based operator used in~\cite{manzoni20}. Hence, we tested a total of 9 combinations of crossover operators and local search variants. Concerning the problem instances, we performed our experiments on Boolean functions of $6 \le n \le 9$ variables. Notice that the number of Boolean functions of $n$ variables is $2^{2^n}$, which means that $n=6$ is the smallest problem instance from where it makes sense to apply metaheuristics, since it is not amenable to exhaustive search.

For the GA, we carried out a preliminary sensitivity analysis by performing small perturbations on the parameters that we adopted in our previous paper~\cite{manzoni20}, to assess if significantly different results would arise. As this did not happen, we sticked to the same GA parameters. In particular, we used a population of $50$ individuals, evolved for a budget of $500\,000$ evaluations, using a steady-state breeding policy with tournament selection of size $t=3$: upon drawing 3 random individuals, the best two are crossed over, and the newly created offspring undergoes mutation with probability 0.7. After calculating the fitness, local search is performed according to the chosen variant, and then the obtained individual replaces the worst one in the tournament. Finally, each experiment (i.e. combination of problem instance, crossover operator and local search policy) was repeated for $30$ independent runs to obtain statistically sound results. To compare two combinations of crossover operator and local search, we adopted the \emph{Mann-Whitney-Wilcoxon} test, with the alternative hypothesis that the corresponding two distributions are not equal, with a significance value $\alpha = 0.05$.

\subsection{Results}
\label{subsec:res}
As expected, the use of local search did not improve significantly the performance of the GA, independently of the underlying combination of crossover and local search policy. The only significant differences arose with the largest instance of $n=9$ variables, where the steepest ascent policy combined with the counter-based and the map-of-ones crossover consistently found functions with a slightly higher nonlinearity of 232 instead of 230 from the other combinations. Since the improvement is anyway too small, we avoid to report the distributions of the best fitness for this case as well.

\begin{figure}[t]
\centering
\begin{subfigure}{.5\textwidth}
\includegraphics[width=1.07\columnwidth]{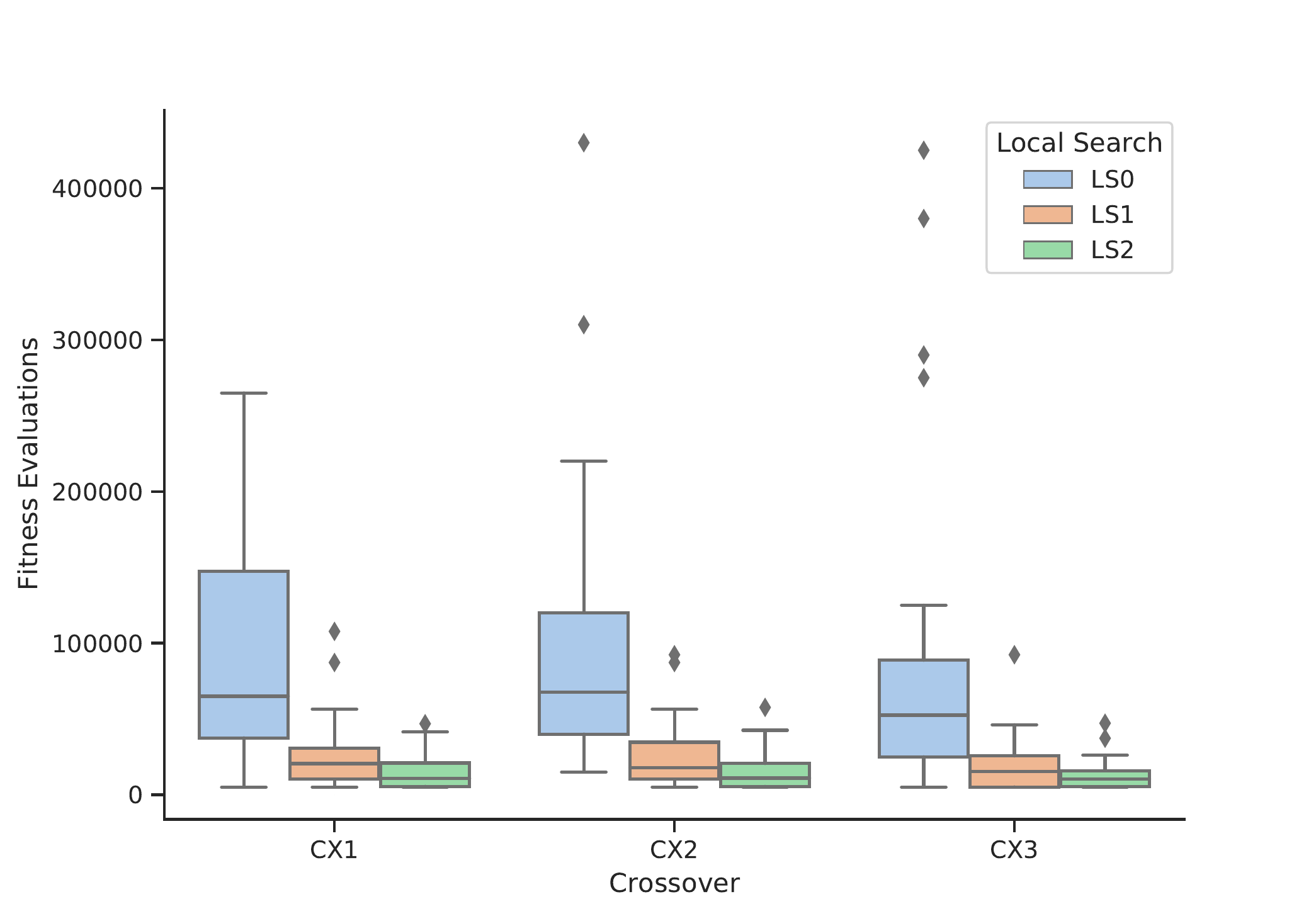}
\caption{$n=6$}
\end{subfigure}%
\begin{subfigure}{.5\textwidth}
\includegraphics[width=1.07\columnwidth]{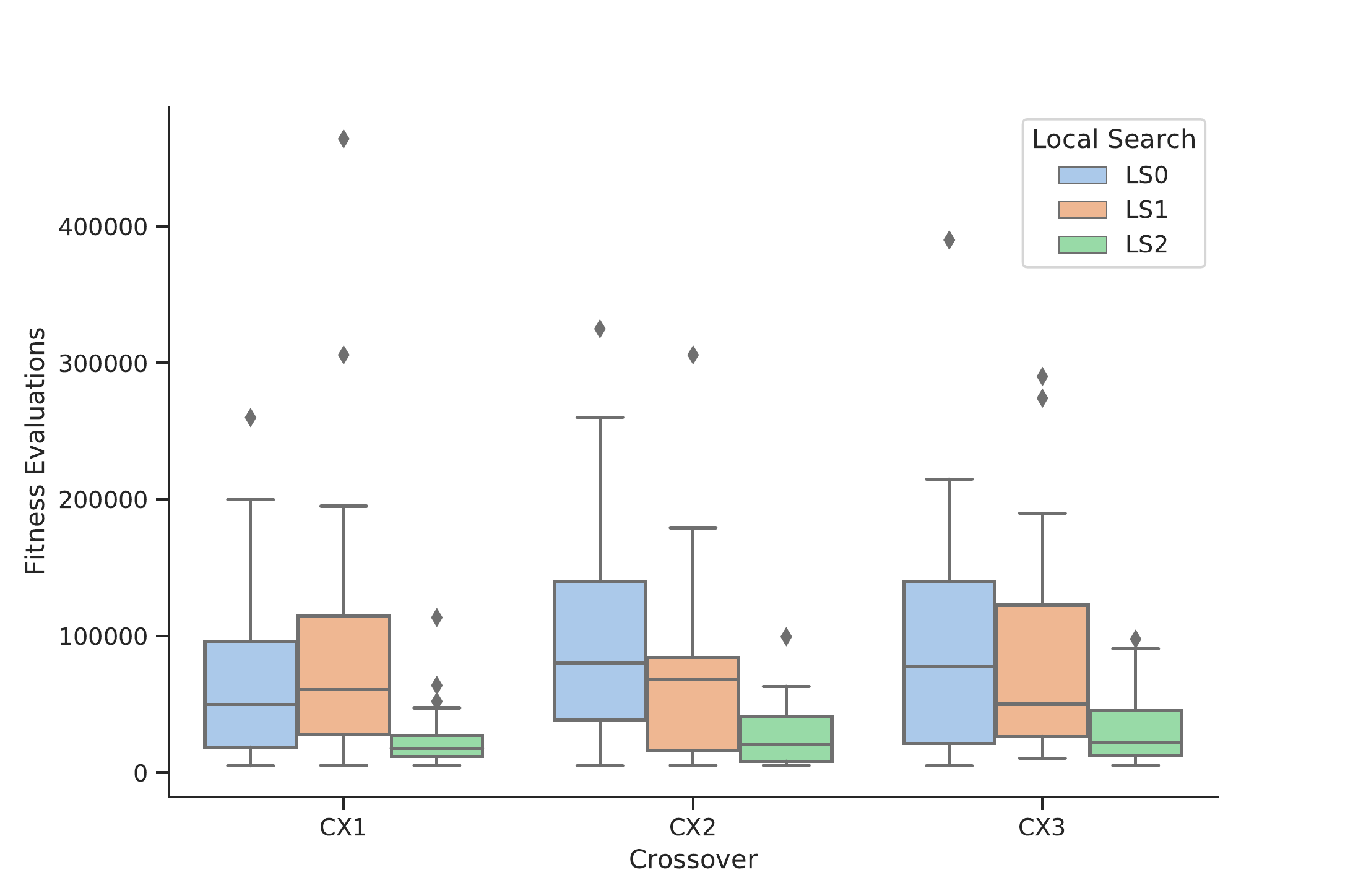}
\caption{$n=7$}
\end{subfigure}

\begin{subfigure}{.5\textwidth}
\includegraphics[width=1.07\columnwidth]{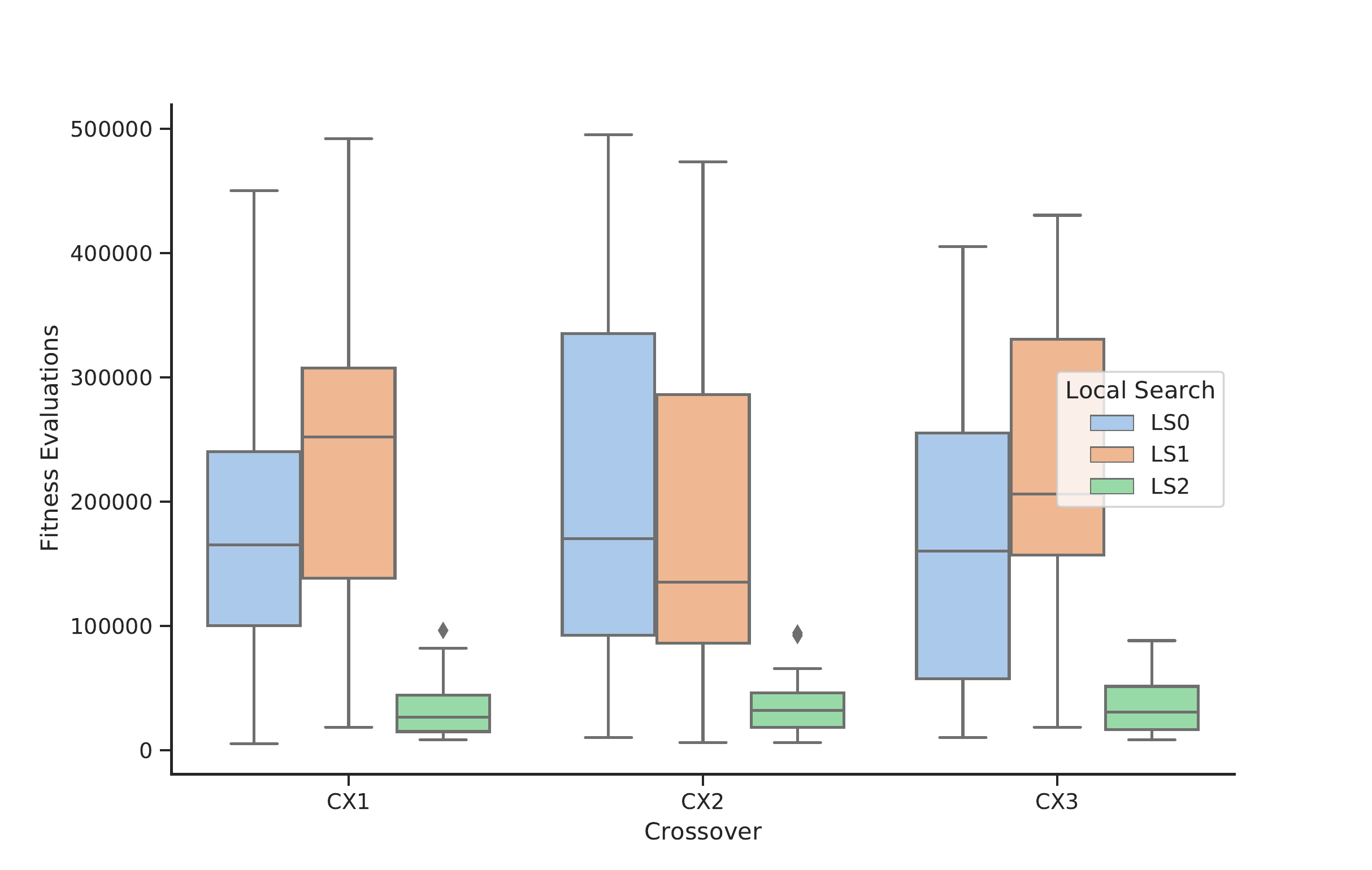}
\caption{$n=8$}
\end{subfigure}%
\begin{subfigure}{.5\textwidth}
\includegraphics[width=1.07\columnwidth]{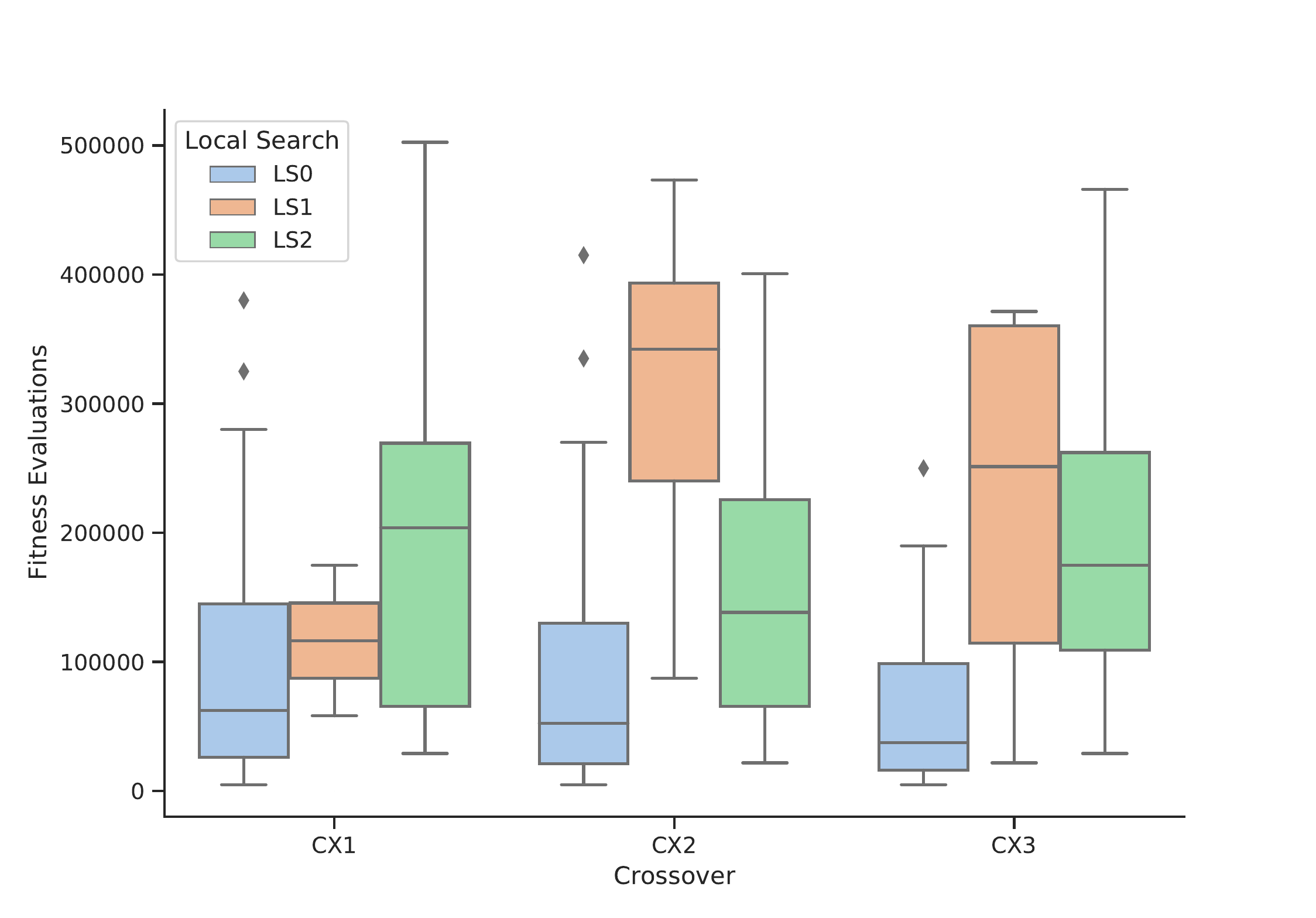}
\caption{$n=9$}
\end{subfigure}
\caption{Boxplots for the distributions of fitness evaluations.}
\label{fig:conv}
\end{figure}

\begin{figure}[t]
     \centering
     \includegraphics[width=0.8\textwidth]{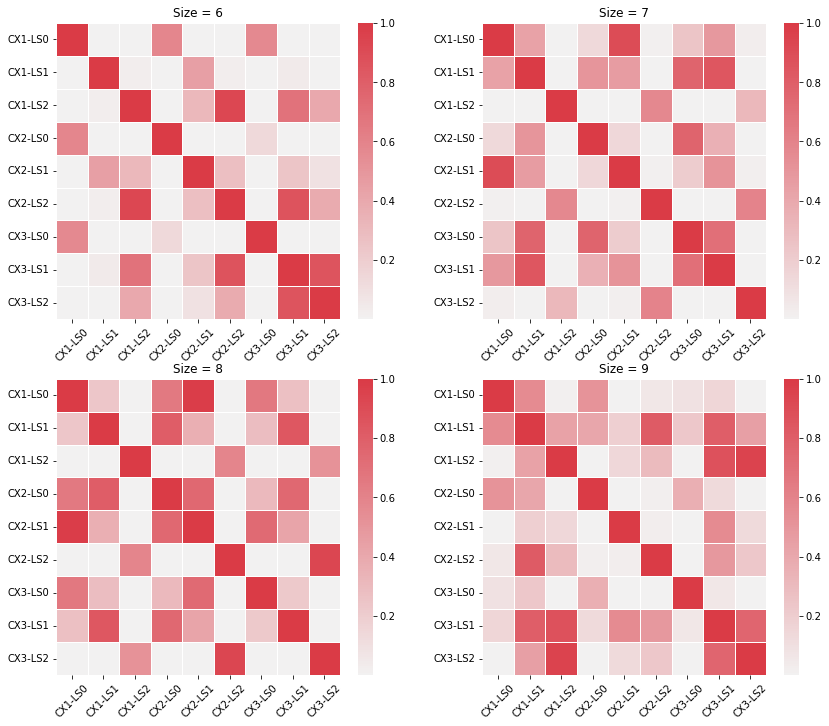}
     \caption{Heatmap of the p-values of the fitness evaluations}
     \label{fig:stat_convergence}
\end{figure}

Figure~\ref{fig:conv} depicts the boxplots for the distributions of the number of fitness evaluations required to reach the best fitness value obtained in each run.
In general, it can be observed that the use of local search does have a substantial effect on the convergence speed of the GA towards a local optimum. This is particular evident in the case of $n=6$ variables for all three crossover operators. For $n=7$ and $n=8$, one can still see from the boxplot that the steepest ascent strategy gives the fastest convergence under all three crossovers, while the situation is less clear for the single-step variant. Looking at the $p$-values heatmaps in Figure~\ref{fig:stat_convergence}, one can indeed see that there are no significant differences between $LS1$ and $LS0$ for all three crossover operators.
The situation seems to be reversed for $n=9$ variables, with the number of fitness evaluations required by the combinations that use the steepest ascent being higher than the variant where no local search is used. Although this finding seems odd at a first glance, it can be easily explained by the remark above on the best fitness. Since for $n=9$ variables the steepest ascent strategy consistently finds Boolean functions with higher nonlinearity than in the basic case, it is reasonable to assume that more fitness evaluations are required to achieve them. 

\begin{figure}[t]
\centering
\begin{subfigure}{.5\textwidth}
\includegraphics[width=1.07\columnwidth]{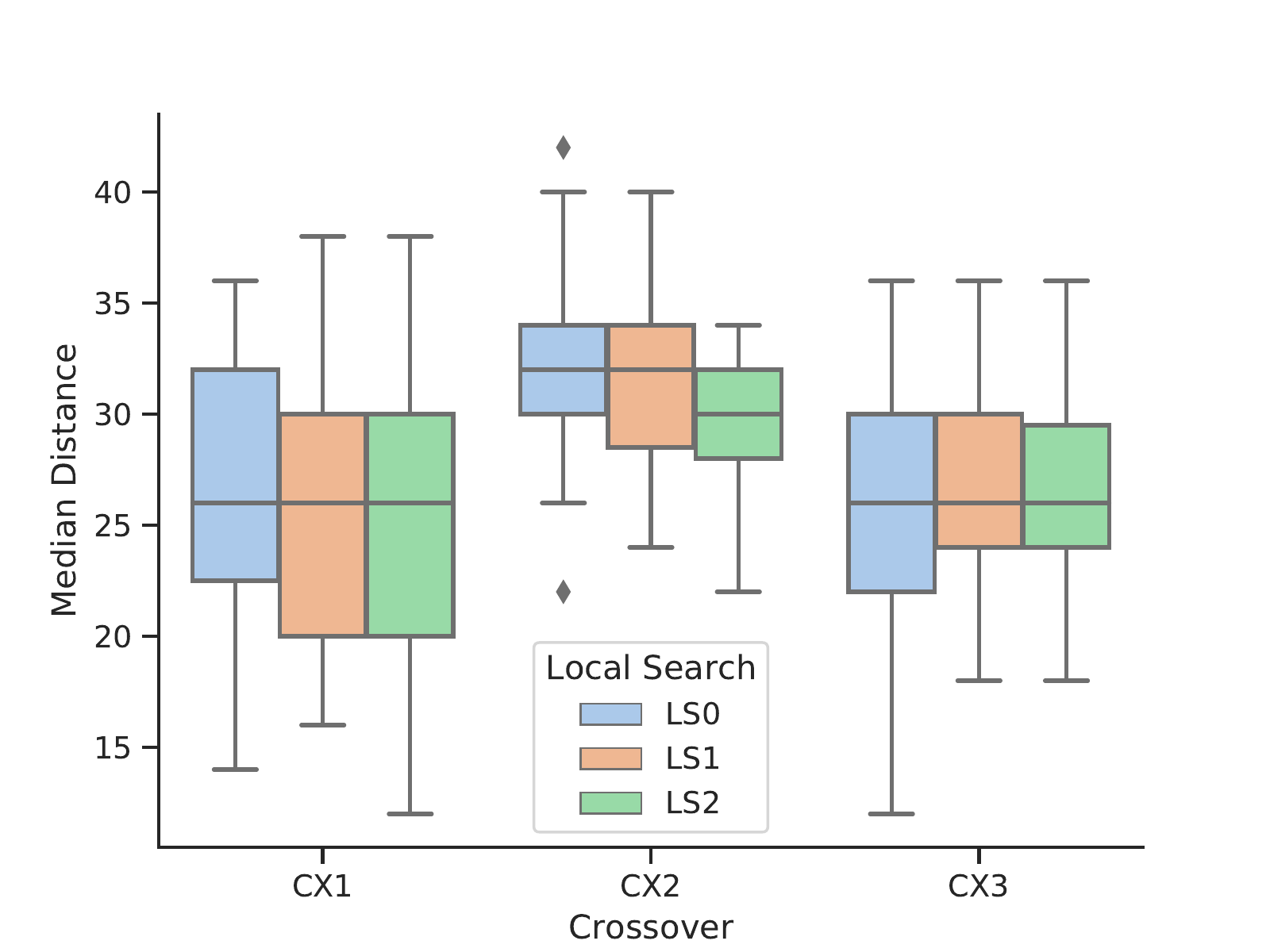}
\caption{$n=6$}
\end{subfigure}%
\begin{subfigure}{.5\textwidth}
\includegraphics[width=1.07\columnwidth]{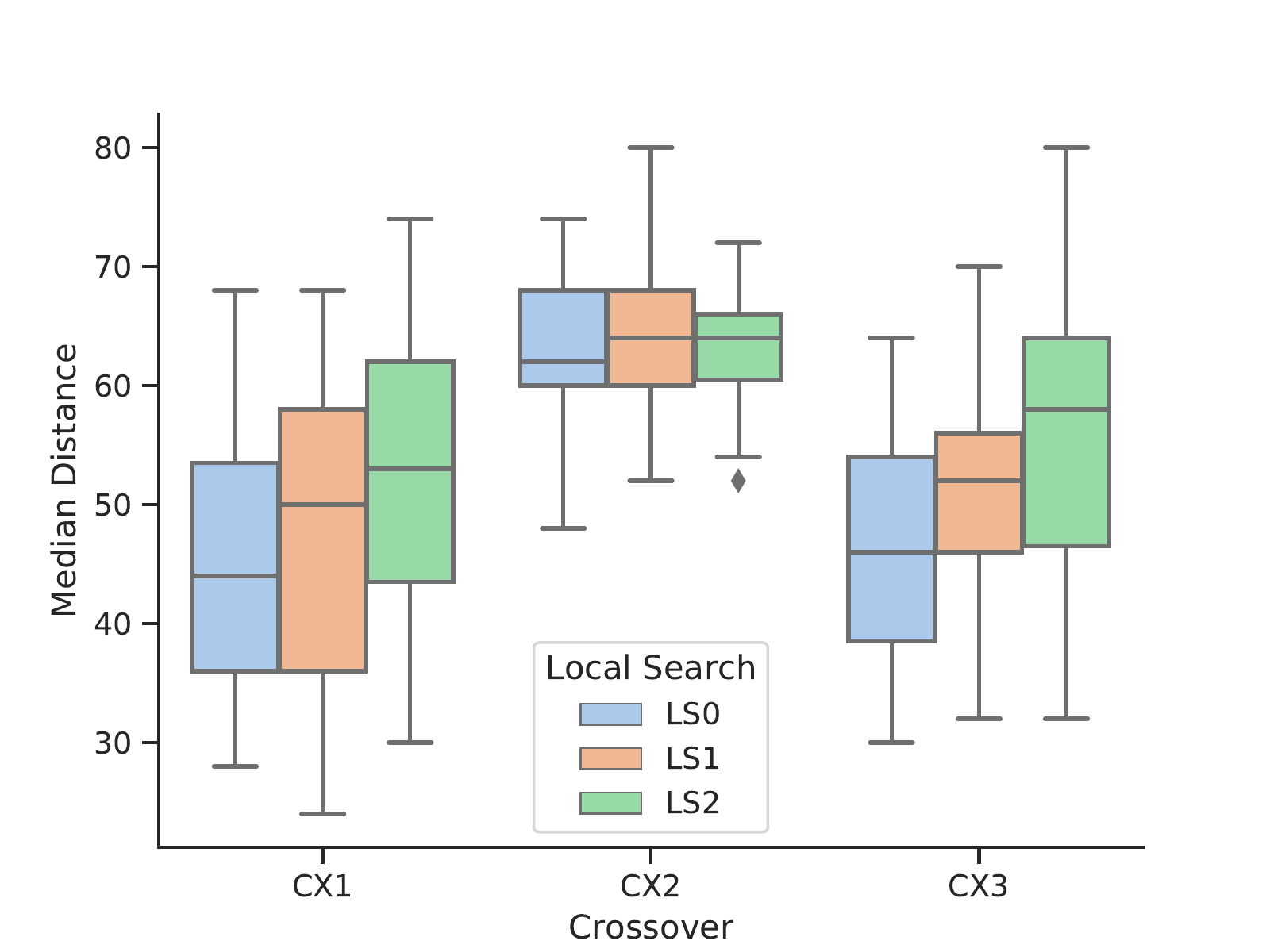}
\caption{$n=7$}
\end{subfigure}

\begin{subfigure}{.5\textwidth}
\includegraphics[width=1.07\columnwidth]{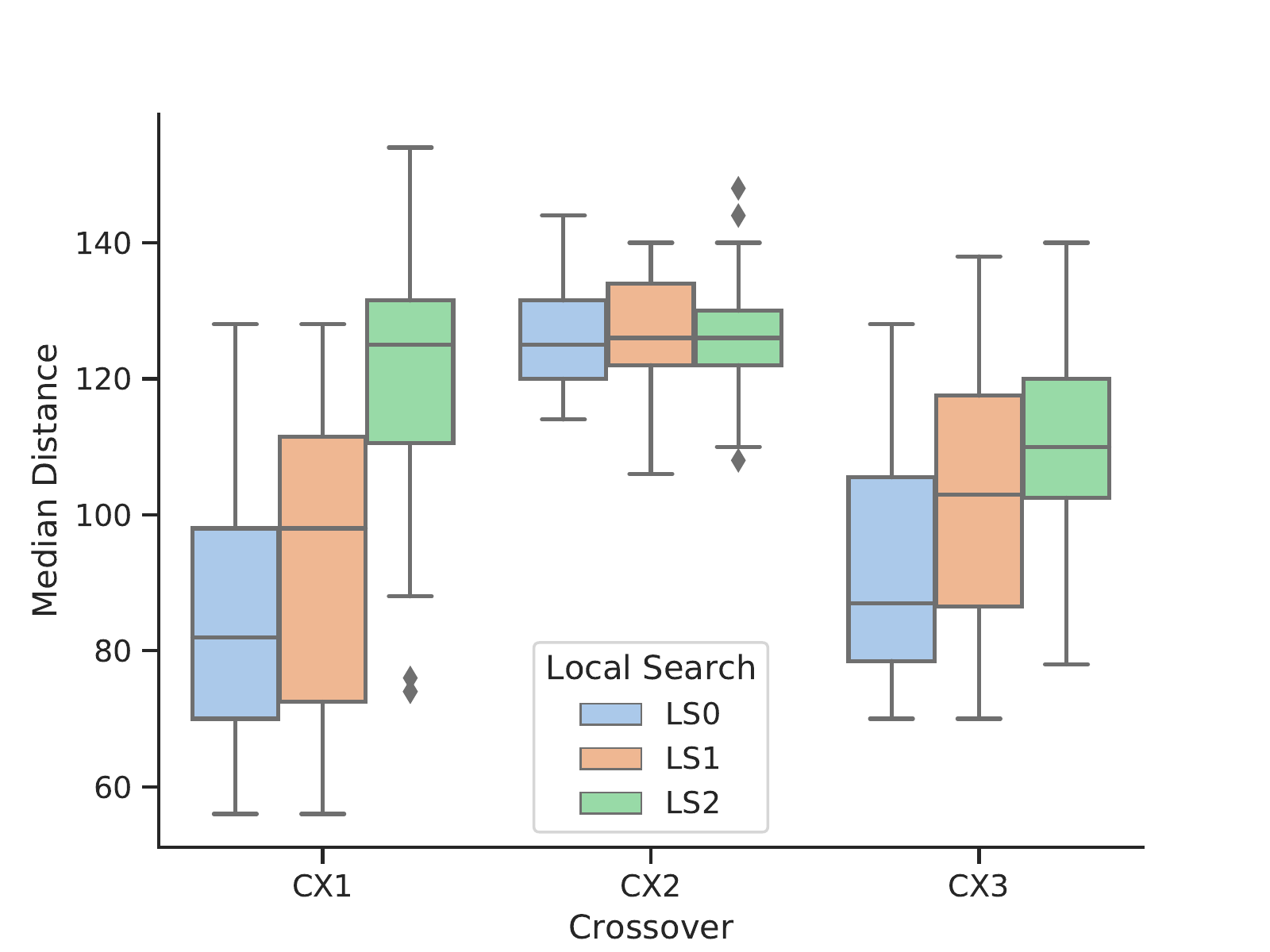}
\caption{$n=8$}
\end{subfigure}%
\begin{subfigure}{.5\textwidth}
\includegraphics[width=1.07\columnwidth]{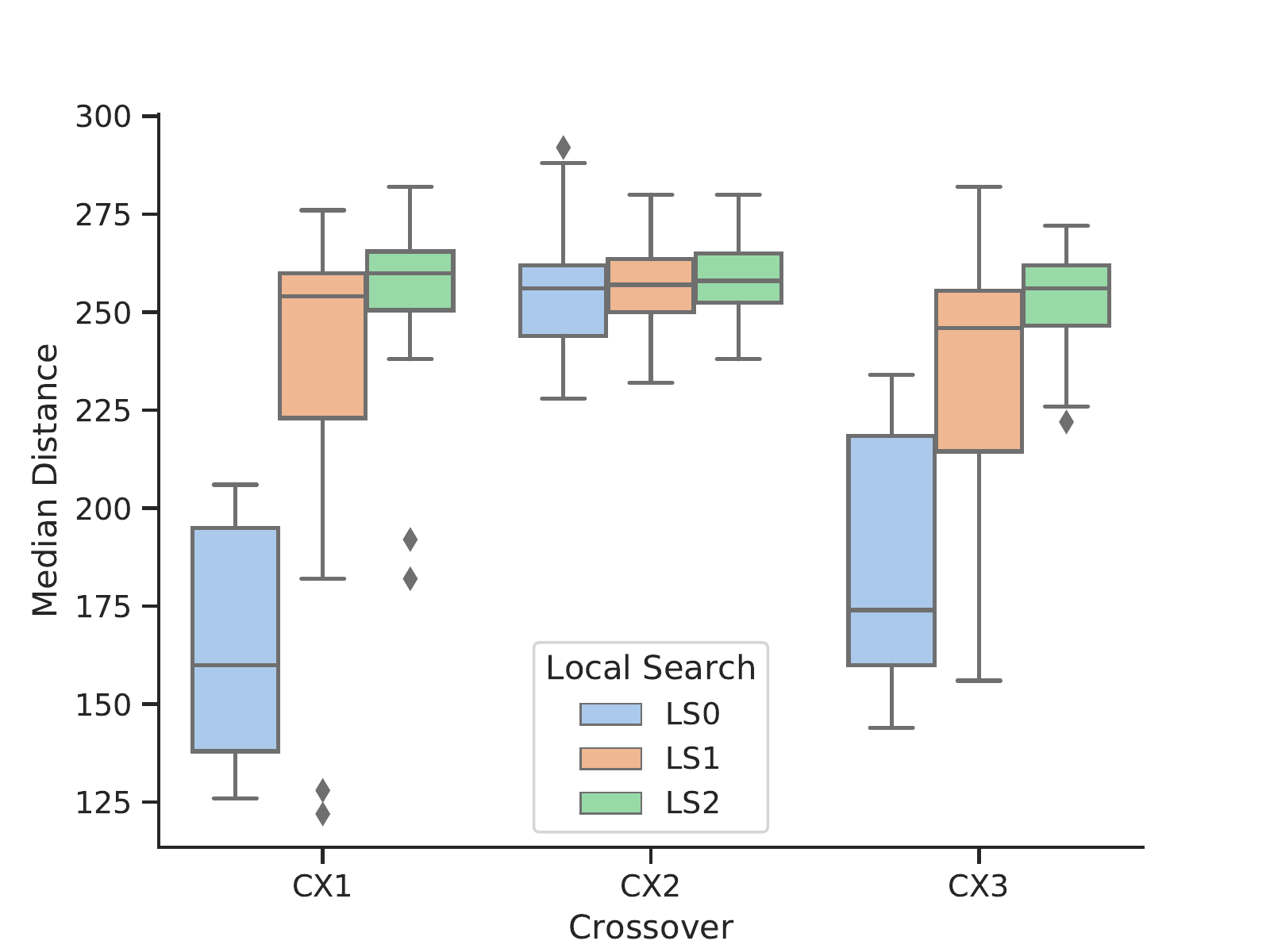}
\caption{$n=9$}
\end{subfigure}
\caption{Boxplots for the distributions of the median distance.}
\label{fig:dist}
\end{figure}

\begin{figure}[t]
     \centering
     \includegraphics[width=0.8\textwidth]{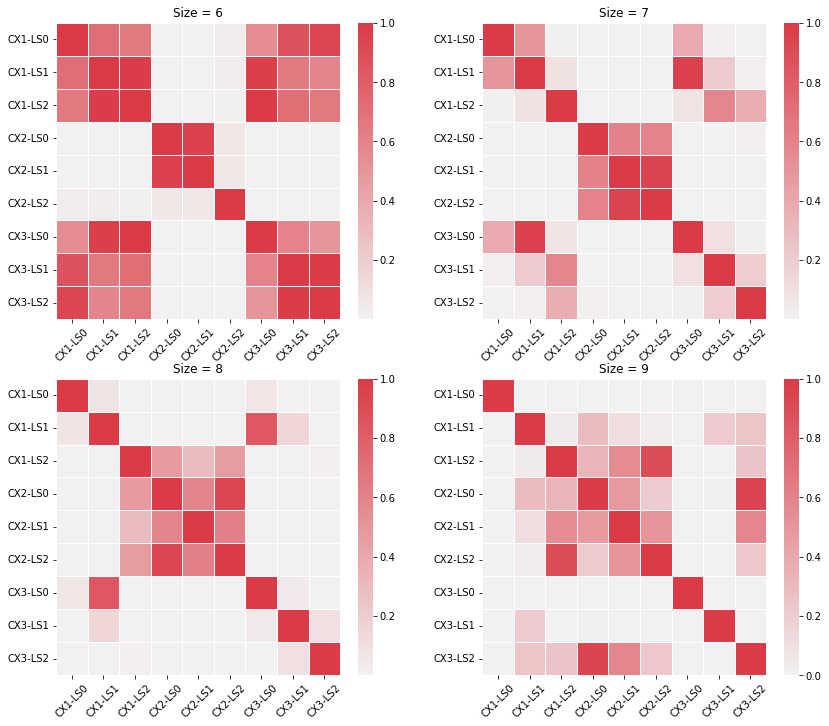}
     \caption{Heatmap of the p-values of the median distance}
     \label{fig:stat_distance}
\end{figure}

To investigate the solutions' diversity, at the end of each run we computed the Hamming distance of each pair of individuals in the population. Figure~\ref{fig:dist} reports the boxplots of the distributions for the median pairwise distance, while Figure~\ref{fig:stat_distance} gives the corresponding $p$-value heatmaps.

The conclusions that one can draw from these results seem counterintuitive: instead of decreasing the population diversity, \emph{the use of local search either does not affect the diversity, or it even increases it in certain cases}. For example, one may see that for $n=6$ there is no difference between the boxplots for each considered crossover, except maybe for $CX2$ where the diversity slightly drops with the steepest ascent policy. This is however not confirmed by the statistical tests, in that no significant differences were observed. By considering bigger instances, one can see that the local search actually starts to play a role in increasing the median distance. This is particularly evident from the boxplots for $n=8$ with the combination of counter-based crossover and steepest ascent, but also for the map-of-ones. The difference becomes even more pronounced for $n=9$ variables, with the steepest ascent obtaining the boxplots with highest median and smallest interquartile range for all three crossovers. This is confirmed by significant differences in the corresponding heatmap. Moreover, in general one can also observe that the zero-length crossover achieves the highest median diversity for all problem instances, independently of the underlying local search policy. Indeed, one can see that the central $3\times 3$ square in each heatmap reports non-significant differences in these cases.
 

\subsection{Discussion}
\label{subsec:disc}
We now attempt to answer the two research questions formulated at the beginning of Section~\ref{sec:exp} in the light of the obtained results.

Concerning {\bfseries RQ1}, the answer seems to be positive: as our initial intuition predicted, the use of local search in general increases the convergence speed of a balanced GA towards a local optimum, independently of the underlying crossover operator. Therefore, although there is no significant improvement in the best fitness (except a slight one for $n=9$ variables), local search allows to reach the current best local optimum more quickly. This is somewhat expected, especially when using a local search step with steepest ascent policy: as each new individual created by GA undergoes local search until a local optimum is reached, the population is quickly filled by candidates that represent local optima, or candidate solutions close to them. Therefore, finding even better local optima by crossing over highly fit individuals in the population might become very unlikely already in the early stages of the optimization process. However, this finding could also indicate that by increasing substantially the fitness budget and the population size of the GA, maybe the best fitness could also improve by employing the steepest ascent local search variant. The rationale is that crossover and mutation could find something better in a large population composed of many local optima obtained through steepest ascent. 

The most interesting finding concerns instead {\bfseries RQ2}. Contrary to our expectations, the use of local search has either little influence over the population diversity, or it even contributes to increase the median Hamming distance among pairs of individuals. This is surprising, as the most natural explanation for the poor performances of balanced GA when compared to other metaheuristics was that the population would converge quickly around a single local optimum, therefore decreasing the population diversity. On the other hand, our experiments confirm that this is not the case, i.e. the final population is composed of many different local optima that are far apart from each other in the search space. A possible explanation of this phenomenon might be related to the shape of the fitness landscape for this particular problem. Indeed, Jakobovic et al.~\cite{jakobovic21} already noticed that the Local Optima Networks (LONs) of generic Boolean functions (i.e., without balancedness constraints) are characterized by a huge number of isolated local optima. Although here we consider a restricted search space, it might still be the case that the resulting fitness landscape has a similar property, since it is a subset of the space of all Boolean functions. In particular, the authors in~\cite{jakobovic21} explained that, to construct a meaningful LON, they had to change the initialization step of their hill climber, so that they could avoid ending up with many isolated local optima. Instead of starting each search trajectory from a completely random point, they employed a \emph{lexicographic sampling}, where each subsequent starting point would be generated in lexicographic order from the first one, which was drawn at random. 

Therefore, a possible insight from the discussion above is that the poor performance of GA in evolving highly nonlinear balanced Boolean functions is not only related to the underlying crossover operators, but also to the method used to initialize the population. Indeed, in our experiments we used a basic initialization step where each individual is generated at random with uniform probability. However, this is exactly what might contribute to cause a high median distance also in the final population, exacerbated by the use of local search, especially in its steepest-ascent version. In future experiments, it would be interesting to test different initialization method, such as the lexicographic sampling mentioned above of~\cite{jakobovic21}, or other methods where the population is created by small random tweaks from a single initial individual.

\section{Conclusions}
\label{sec:outro}
In this work, we investigated the effect of a local search step combined with balanced GA to evolve highly nonlinear balanced Boolean functions. The motivation was to analyze the possible causes of the poor performances of balanced GA on this particular optimization problem, when compared to other metaheuristics such as GP. To this end, we set up our investigation by adding to the GA with balanced crossovers proposed in our previous paper~\cite{manzoni20} a local search strategy originally devised by Millan et al.~\cite{millan99}. We investigated three variants, namely no local search, single-step local search, and steepest-ascent local search, and applied it to the optimization of Boolean functions of $6 \le n \le 9$ variables. The investigation was centered around two main research questions: the first one concerned whether the use of local search increased the convergence speed of a balanced GA toward a local optimum. The second question asked if local search decreases the population diversity, as measured by the median pairwise Hamming distance between individuals. While our results answered affirmatively the first question as expected, the answer to the second question surprisingly turned out to be negative. In particular, local search either does not affect or even increases the median distance in the population. We discussed this finding by referring to a recent work on the fitness landscapes of Boolean functions~\cite{jakobovic21}, in the form of Local Optima Networks. In particular, the main insight gained from this discussion is that the poor performance of balanced GA might be connected to the initialization method of the population, which right now generates each individual independently with uniform probability.

Future experiments should consider other types of initialization as well, such as a random walk from a single initial individual, or lexicographic generation. Beside this, several other directions for future research remained to be explored on the subject. Perhaps the most interesting one, after the finding of this paper, involves the analysis of the fitness landscape for the particular search space of balanced Boolean functions. Indeed, the analysis of Local Optima Networks in~\cite{jakobovic21} considered the space of all Boolean functions, with no balancedness constraints. Therefore, it would be interesting to repeat the analysis for balanced functions, to see if similar properties like many isolated local optima still emerge. Further, we believe that it would be interesting to augment GA with local search also for other optimization problems that require balanced representations, such as the construction of bent functions and orthogonal arrays already considered in our previous paper~\cite{manzoni20}.

\bibliographystyle{abbrv}
\bibliography{bibliography}

\end{document}